\documentclass[letterpaper, 10 pt, conference]{ieeeconf}  

\IEEEoverridecommandlockouts                              
\usepackage[utf8]{inputenc}
\usepackage[T1]{fontenc}
\usepackage{hyperref}
\usepackage{graphics} 
\usepackage{epsfig} 
\usepackage{mathptmx} 
\usepackage{amsmath} 
\usepackage{amssymb}  
\usepackage{booktabs}
\usepackage{multirow}
\usepackage{microtype}
\usepackage{mathrsfs}
\usepackage{bbm}
\usepackage{calrsfs}
\usepackage[normalem]{ulem}
\usepackage{threeparttable}
\useunder{\uline}{\ul}{}
\DeclareMathAlphabet{\pazocal}{OMS}{zplm}{m}{n}

\title{\LARGE \bf
RL-BioAug: Label-Efficient Reinforcement Learning \\for Self-Supervised EEG Representation Learning}

\author{Cheol-Hui Lee$^{1}$, Hwa-Yeon Lee$^{1}$, and Dong-Joo Kim$^{1,2,3}$
\thanks{$^{1}$The authors are with the Department of Brain and Cognitive Engineering, Korea University, Seoul, South Korea.}%
\thanks{$^{2}$Dong-Joo Kim is with the Interdisciplinary Program in Precision Public Health, Korea University, Seoul, South Korea.}%
\thanks{$^{3}$Dong-Joo Kim is with the Department of Neurology, Korea University College of Medicine, Seoul, South Korea (Corresponding author: {\tt\small dongjookim@korea.ac.kr})}
}

\begin{document}

\maketitle
\thispagestyle{empty}
\pagestyle{empty}

\begin{abstract}

The quality of data augmentation serves as a critical determinant for the performance of contrastive learning in EEG tasks. Although this paradigm is promising for utilizing unlabeled data, static or random augmentation strategies often fail to preserve intrinsic information due to the non-stationarity of EEG signals where statistical properties change over time. To address this, we propose RL-BioAug, a framework that leverages a label-efficient reinforcement learning (RL) agent to autonomously determine optimal augmentation policies. While utilizing only a minimal fraction (10\%) of labeled data to guide the agent's policy, our method enables the encoder to learn robust representations in a strictly self-supervised manner. Experimental results demonstrate that RL-BioAug significantly outperforms the random selection strategy, achieving substantial improvements of 9.69\% and 8.80\% in Macro-F1 score on the Sleep-EDFX and CHB-MIT datasets, respectively. Notably, this agent mainly chose optimal strategies for each task---for example, Time Masking with a 62\% probability for sleep stage classification and Crop \& Resize with a 77\% probability for seizure detection. Our framework suggests its potential to replace conventional heuristic-based augmentations and establish a new autonomous paradigm for data augmentation. The source code is available at \href{https://github.com/dlcjfgmlnasa/RL-BioAug}{https://github.com/dlcjfgmlnasa/RL-BioAug}.


\end{abstract}

\section{INTRODUCTION}
Healthcare and EEG analysis have achieved remarkable performance improvements due to rapid advancements in deep learning \cite{roy2019deep, craik2019deep}. Despite this progress, many studies rely on supervised learning, fundamentally limited by the need for large-scale datasets labeled by experts \cite{banville2021uncovering}. The resulting substantial costs and time requirements serve as a primary bottleneck for the practical deployment of deep learning models in clinical settings \cite{kostas2021bendr}.

To address this challenge, self-supervised learning (SSL)—particularly contrastive learning—has gained attention for its ability to learn representations without labels in EEG analysis \cite{lee2024neuronet, eldele2021time}. The core principle of contrastive learning is to maximize the similarity between positive pairs generated by data augmentation \cite{chen2020simple}. Consequently, the choice of augmentation technique is a critical factor that determines the invariance that the model has to learn \cite{xiao2020should}. While the standardized strategies are well-established in computer vision, a unified methodology lacks in EEG analysis due to the complex nature of the signals \cite{rommel2022data}. Notably, data augmentation is not just an auxiliary tool for generalization as in supervised learning; it is a fundamental component in contrastive learning that shapes that quality of the representations \cite{chen2020simple, tian2020makes}.

However, the non-stationarity of EEG signals poses a significant challenge in defining an optimal augmentation strategy \cite{kaplan2005nonstationary}. Data augmentation hinges on a delicate balance between inducing sufficient diversity and preserving intrinsic integrity; yet, a static and uniform approach that neglects signal states is liable to distort the inherent patterns of data in EEG analysis tasks (e.g., sleep stage classification, seizure detection). For instance, high-intensity noises are relatively tolerable in the Wake state with frequent EMG artifacts \cite{berry2017aasm}. In contrast, REM sleep or Seizure intervals are fragile to the same intensity where preserving subtle waveform variations is critical \cite{berry2017aasm, boonyakitanont2020review}. These strong distortions can lead to the loss of intrinsic features, ultimately degrading the quality of representation learning \cite{iwana2021empirical, eldele2021time}. Therefore, an adaptive strategy that dynamically adjusts augmentation types based on the signal state is essential.

We validated the effectiveness of the proposed method using two EEG benchmarks: the Sleep-EDFX \cite{kemp2000analysis} and CHB-MIT \cite{shoeb2009application} datasets. Experimental results confirmed that our framework achieves superior classification performance compared to random or fixed augmentation strategies.

\section{METHODS}

\begin{figure*}[!t]
    \centerline{\includegraphics[width=0.9\textwidth]{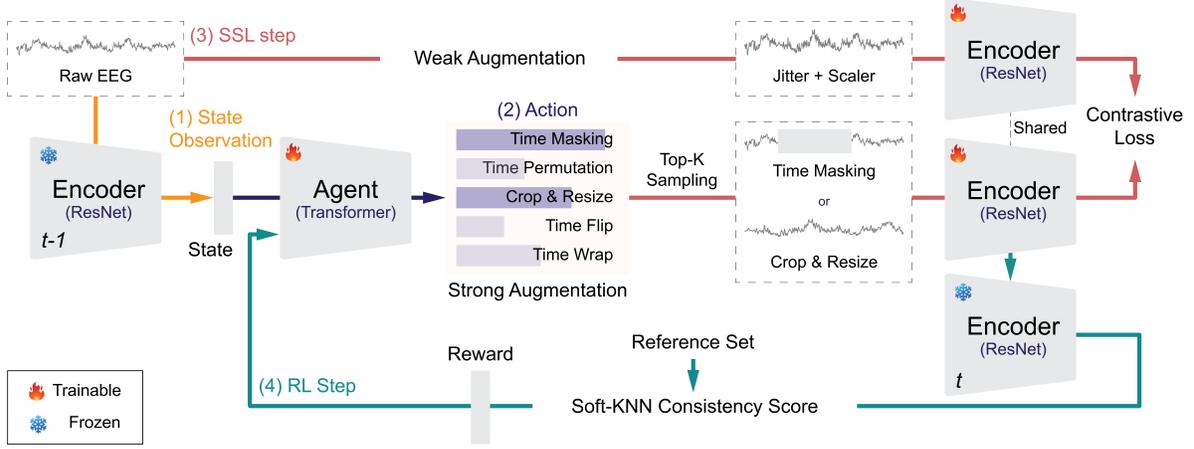}}
    \caption{\textbf{Overview of the RL-BioAug Framework.} The process consists of four distinct phases: \textit{(1) State Observation}, where the state is extracted via a frozen encoder; \textit{(2) Action}, where the agent selects an optimal strong augmentation; \textit{(3) SSL Step}, which updates the encoder using contrastive loss between weak and strong views; and \textit{(4) RL Step}, which evaluates the representation quality using the Soft-KNN consistency score on a reference set to reward the agent.}
    \label{fig:figure1}
\end{figure*}

In this study, we propose RL-BioAug, a framework that simultaneously optimizes contrastive learning and data augmentation policies for EEG signals. The framework allows the encoder and the agent to jointly optimize their parameters, driving the cooperative learning process toward the most effective data augmentation strategy.

\subsection{Cooperative Optimization Loop}

RL-BioAug operates on a cyclic structure at every training step in the following sequence: \textbf{State Observation} $\to$ \textbf{Action} $\to$ \textbf{SSL Step}: \textit{encoder update via contrastive loss} $\to$ \textbf{RL Step}: \textit{agent update via policy gradient}.

\subsubsection{Non-causal Environments and Context-based Decision} RL is generally premised on causal assumption that actions of an agent alter the distribution of future states. 
In contrast, applying augmentation to a specific sample in our environment does not influence the physical properties of subsequent, independent samples. 
Given this non-causal nature, we designed the agent to focus solely on analyzing the immediate context rather than formulating long-term future plans.

\subsubsection{Reinforcement Learning Problem Formulation}
The key components of this framework are defined as follows:

\begin{itemize}
    \item \textbf{State} ($s_t$) -- A state is defined as the embedding vector extracted from the freeze encoder ($f_\theta$) at the current time step. This vector encapsulates the semantic information of the original signal and provides it to the agent. We employed ResNet18-1d~\cite{wang2017time} as the encoder.
    
    \item \textbf{Action} ($a_t$) -- The agent selects and executes the optimal strong augmentation from a discrete action space ($\mathbb{A}$), which consists of five pre-defined transformations: \textit{Time Masking}, \textit{Time Permutation}, \textit{Crop \& Resize}, \textit{Time Flip}, and \textit{Time Warp}.
    
    \item \textbf{Reward} ($r_t$) -- To quantitatively assess whether the selected action contributes to the improvement of the encoder's representation capability, we utilize the Soft-KNN consistency score.
\end{itemize}

The detailed workflow of RL-BioAug is organized as follows: Section \ref{sec:agent} describes the agent that performs actions. Sections \ref{sec:ssl_step} and \ref{sec:rl_step} present the overall training process including SSL and RL steps.

\subsection{Agent: Transformer-based Policy Network} \label{sec:agent}
The agent selects its augmentation policy by leveraging past augmentation history and reward feedback. To achieve this, we employ a Transformer-based policy network ($\pi_\varphi$), which is renowned for its efficacy in processing sequential data.

\begin{itemize}
    \item \textbf{Feature Fusion} -- The input to the agent comprises the current state representation ($s_t$) and the sequence of past $K$ actions and rewards (i.e., $\{a_{t-K}, \dots, a_{t-1}\}$ and $\{r_{t-K}, \dots, r_{t-1}\}$). These components are processed through an FC layer to be transformed into an integrated context vector.
    \item \textbf{Sequence Analysis} -- Positional embeddings are added to the integrated sequence before it is fed into the Transformer. The self-attention mechanism captures the historical context of policy-reward dynamics to infer the current representation state.
    \item \textbf{Action Selection} -- The representation vector at the final time step $t$ is passed through an FC layer to generate an action probability distribution. Finally, the action is selected via Top-K Sampling, which effectively balances exploration and stability.
\end{itemize}

\subsection{Self-Supervised Learning Step: Encoder Update via Contrastive Loss} \label{sec:ssl_step}

The selected action is immediately applied to the original signal $x$ to generate a strongly augmented view $x_{strong}$. This view is then paired with a weakly augmented view $x_{weak}$ for contrastive learning. Notably, the encoder parameters are instantaneously updated based on the view generated by the agent. We selected SimCLR \cite{chen2020simple} as our contrastive learning framework, given its high sensitivity to the quality of data augmentation \cite{chen2020simple} \cite{tian2020makes}. 
\begin{gather} \label{eq:eq_1}
    \theta \leftarrow \theta - \eta_{\text{enc}} \nabla_\theta \pazocal{L}_{\text{InfoNCE}}(f_\theta(x_{\text{weak}}), f_\theta(x_{\text{strong}}))
\end{gather}
\begin{gather} \label{eq:eq_2}
    \pazocal{L}_{\text{InfoNCE}} = -\log \frac{\exp(\text{sim}(z_{\text{weak}}, z_{\text{strong}})/\tau)}{\sum_{k=1}^{2N} \mathbb{1}_{[k \neq i]} \exp(\text{sim}(z_{\text{weak}}, z_k)/\tau)}
\end{gather} 
here, Eq.~(\ref{eq:eq_1}) presents the update rule with learning rate $\eta_{\text{enc}}$. With the encoded representations $z = f_\theta(x)$, Eq.~(\ref{eq:eq_2}) aligns positive pairs ($z_{\text{weak}}, z_{\text{strong}}$) while repelling negatives ($z_k$) scaled by temperature $\tau$.

This process ensures that the agent's action induces structural changes in the encoder's representation space, providing a direct causal basis for the reward calculation.

\subsection{Reinforcement Learning Step: Agent Update via Policy Gradient} \label{sec:rl_step}
\subsubsection{Reward Definition via Soft-KNN}
We utilize Soft-KNN consistency score to quantitatively assess whether the agent’s selected action contributes to improving the encoder’s representation capability. This score measures how tightly the embeddings of the current batch form clusters with same-class neighbors from a distinct reference set in the embedding space.

Based on the cosine similarity $sim(z_i, z_j)$ between the embedding of the current sample $z_i$ and the embedding of a sample $z_j$ in the reference set, the probability $P(y_i = c | z_i)$ that sample $i$ belongs to a specific class $c$ is derived as a softmax-weighted sum of the labels of the top-$K$ nearest neighbors:
\begin{equation}
P(y_i = c | z_i) = \frac{\sum_{j \in \pazocal{N}_i} \exp(sim(z_i, z_j) / \tau) \cdot \mathbb{1}(y_j = c)}{\sum_{k \in \pazocal{N}_i} \exp(sim(z_i, z_k) / \tau)}
\end{equation}
Here, $\pazocal{N}_i$ denotes the set of $K$ nearest neighbors of $z_i$ within the reference set, and $\mathbb{1}(\cdot)$ is an indicator function that returns 1 if the condition is met and 0 otherwise. The final reward $r_t$ received by the agent is defined as the probability value corresponding to the actual ground-truth label ($y_{true}$) of the sample, providing a dense and continuous feedback signal between 0 and 1:

\begin{equation}
r_t = P(y_i = y_{true} | z_i) = \frac{\sum_{j \in \mathcal{N}_i} \exp(sim(z_i, z_j) / \tau) \cdot \mathbb{1}(y_j = y_{true})}{\sum_{k \in \mathcal{N}_i} \exp(sim(z_i, z_k) / \tau)}
\end{equation}

\subsubsection{REINFORCE++ Optimization} To ensure the stable convergence of the policy network in noisy EEG-signal environments, we adopt the REINFORCE++ algorithm~\cite{hu2025reinforce++}.

\paragraph{Variance Reduction via Advantage}
To suppress the variance of the policy gradient, we utilize the advantage function $A_t$. We define the baseline $b_t$ as the average reward across the mini-batch $B$ and subtract it from the observed reward $r_t$ to compute the advantage:
\begin{equation}
    A_t = r_t - b_t, \quad \text{where} \quad b_t = \frac{1}{|B|} \sum_{i=1}^{|B|} r_i
\end{equation}

\paragraph{Entropy Regularization}
To prevent the agent from prematurely converging to a deterministic policy that relies on a limited set of augmentations, we incorporate an entropy regularization term. We add the entropy of the policy distribution $H(\pi_\varphi)$ to the objective function with a coefficient $\gamma = 0.1$. This explicit regularization encourages the agent to maintain high entropy, thereby promoting active exploration of diverse augmentation strategies.

\paragraph{Final Objective Function}
To define the optimization process, we apply a gradient descent step to update the policy network parameters $\varphi$ with a learning rate $\eta_{\text{agent}}$:
\begin{equation} \label{eq:policy_update}
    \varphi \leftarrow \varphi - \eta_{\text{agent}} \, \nabla_\varphi \pazocal{L}_{\text{Policy}}(\varphi)
\end{equation}
The objective function $\pazocal{L}_{\text{Policy}}(\varphi)$ is formulated to maximize the expected advantage while maintaining policy entropy:
\begin{equation} \label{eq:policy_loss}
    \pazocal{L}_{\text{Policy}}(\varphi) = -\mathbb{E} \Big[ \log \pi_\varphi(a_t|s_t) \, A_t \Big] - \beta \gamma H(\pi_\varphi)
\end{equation}
here, the policy entropy term $H(\pi_\varphi)$, which encourages exploration over the action space $\mathbb{A}$, is explicitly defined as:
\begin{equation} \label{eq:entropy}
    H(\pi_\varphi) = - \sum_{a \in \mathbb{A}} \pi_\varphi(a|s_t) \log \pi_\varphi(a|s_t)
\end{equation}
Finally, $\beta$ serves as a decaying factor that controls the trade-off between exploration and exploitation. As training progresses, $\beta$ is gradually reduced to allow the agent to shift its focus from exploring new strategies to exploiting the learned optimal policy.

\begin{table*}[!t]
\centering
\begin{threeparttable}
\caption{Comparison of augmentation strategies on Sleep-EDFX and CHB-MIT datasets.}
\label{tab:table1}
\small 
\setlength{\tabcolsep}{18pt} 
\begin{tabular}{cc cccc}
\toprule
\multirow{2.5}{*}{\textit{Method}} & \multirow{2.5}{*}{\textit{\shortstack{Augmentation\\Strategy}}} & \multicolumn{2}{c}{Sleep-EDFX} & \multicolumn{2}{c}{CHB-MIT} \\
\cmidrule(lr){3-4} \cmidrule(lr){5-6}
 & & B-ACC & MF1 & B-ACC & MF1 \\
\midrule
\multirow{5}{*}{Single} & Time Masking & \uline{65.52} & \uline{61.83} & 68.50 & 63.15 \\
 & Time Permutation & 65.33 & 60.82 & 67.45 & 61.80 \\
 & Crop \& Resize & 64.63 & 60.40 & \uline{69.12} & \uline{64.25} \\
 & Time Flip & 64.22 & 59.52 & 66.82 & 60.40 \\
 & Time Warp & 62.49 & 57.80 & 65.10 & 58.95 \\
\midrule
\multirow{2}{*}{Composite} & Random Selection & 64.48 & 59.86 & 68.05 & 62.70 \\
 & \textbf{RL-BioAug (Ours)} & \textbf{72.72} & \textbf{69.55} & \textbf{76.24} & \textbf{71.50} \\
\bottomrule
\end{tabular}
\begin{tablenotes}
    \footnotesize
    \item - The \textbf{best results} in each column are shown in bold, while the \uline{second-best} results are underlined.
\end{tablenotes}
\end{threeparttable}
\end{table*}

\section{EXPERIMENTS}
\subsection{Dataset Description}
We utilized two EEG datasets with distinct characteristics to evaluate the generalizability and robustness of our proposed framework. For both datasets, we applied a 0.5--40 Hz band-pass filter and performed z-normalization.

\begin{itemize}
    \item \textbf{Sleep-EDFX} \cite{kemp2000analysis} -- We employed the Sleep-Cassette subset of the 
    Sleep-EDFX dataset for sleep stage classification and used the single Fpz-Cz channel. In accordance with The American Academy of Sleep Medicine standards, we merged the N3 and N4 stages to reconfigure the data into five classes: Wake, N1, N2, N3, and REM.
    
    \item \textbf{CHB-MIT} \cite{shoeb2009application} -- We utilized the CHB-MIT Scalp EEG dataset for epileptic seizure detection. The single Fp1-F7 channel was selected. The signals were segmented into 4-second windows, and the labels were reconstructed for binary classification: Seizure and Background.
\end{itemize}

\begin{figure}[b]
    \centerline{\includegraphics[width=1\columnwidth]{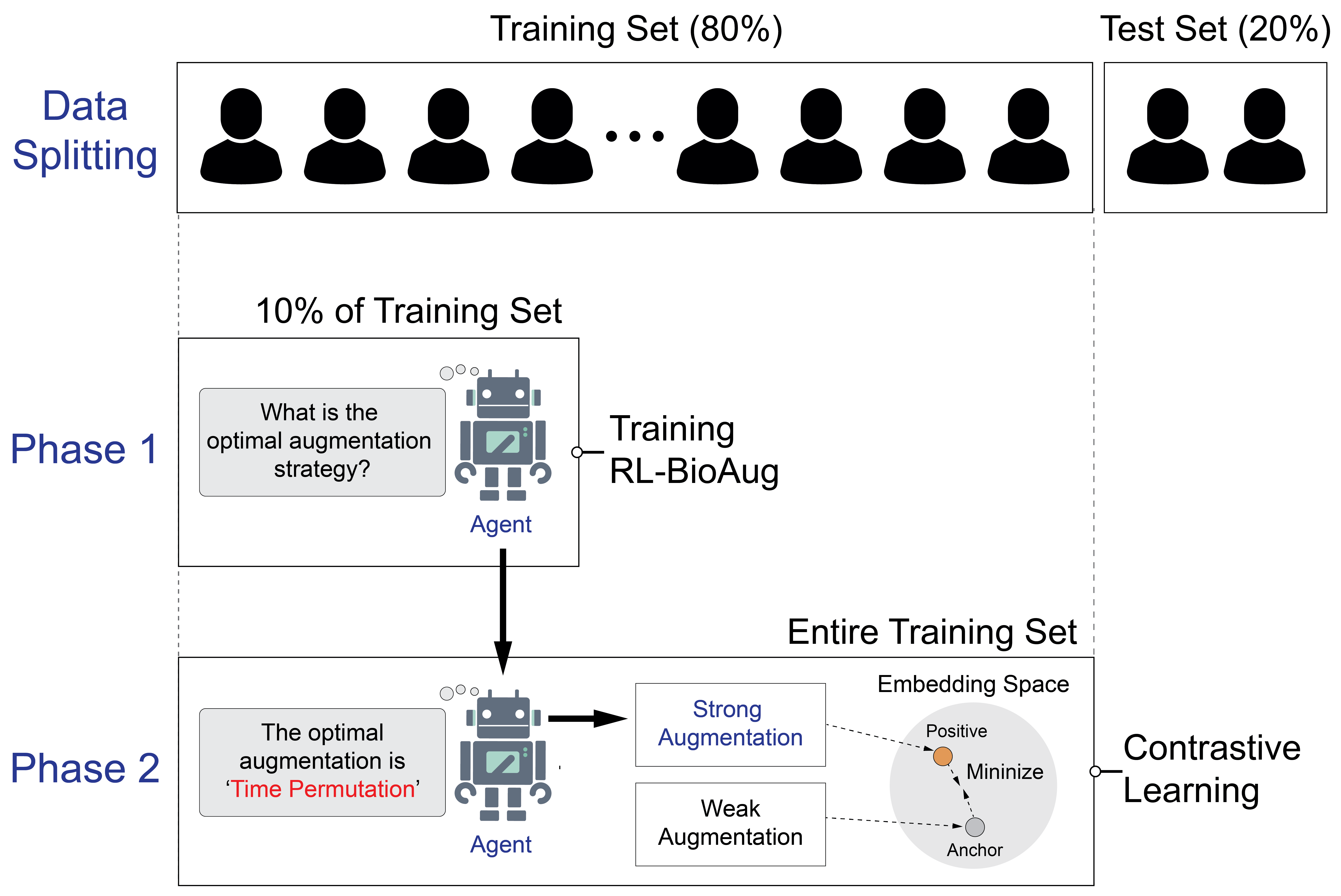}}
    \caption{Experiment Protocol.}
    \label{fig:figure2}
\end{figure}

\subsection{Experiment Protocol}
We designed an experimental protocol to validate the effectiveness of RL-BioAug.

\begin{enumerate}
    \item \textbf{Data Splitting (Subject-independent)} --
    To evaluate robust generalization performance, we adopted a subject-independent split strategy, strictly partitioning training and testing datasets by subject. 80\% of the subjects were allocated for training, while the remaining 20\% were allocated for testing.

    \item \textbf{Phase 1: Training RL-BioAug} --
    In Phase 1, the RL agent learns the optimal data augmentation policy. Given that the reward function (Soft-KNN Consistency Score) relies on ground-truth labels, we trained the agent using only a small fraction (10\%) of the labeled data from the training set. Additionally, we utilized a portion of this subset as the \textit{Reference set} to compute the reward. This demonstrates that meaningful policy learning is achievable with minimal supervision.

    \item \textbf{Phase 2: Contrastive Learning} --
    In Phase 2, the encoder is trained while the pre-trained agent remains frozen. We utilized the entire training dataset for SSL, completely excluding labels. This design ensures that the model learns representation solely through the augmented views generated by the agent.
\end{enumerate}

\subsection{Data Augmentation} 
\subsubsection{Strong Augmentation} \label{sec:strong_augmentation}
The RL-BioAug agent selects one technique from a pool of strong augmentations that significantly alter the temporal or structural characteristics of the signals. This approach encourages the encoder to maintain invariance against challenging views during contrastive learning. The details of the five strong augmentation techniques are as follows:

\begin{itemize}
    \item \textbf{Time Masking} masks arbitrary time intervals with zeros. It simulates electrode artifacts (e.g., pop-outs) and compels the model to predict the missing intervals based on the surrounding context.
    
    \item \textbf{Time Permutation} segments the signal into multiple intervals and randomly shuffles their order. It encourages the model to learn the global structure and sequential information of the data.
    
    \item \textbf{Crop \& Resize} crops a local temporal region and resizes it to the original length. It prevents reliance on global patterns and forces the model to focus on detailed waveform features.
    
    \item \textbf{Time Flip} reverses the time axis. It induces the model to learn features that remain robust even when the temporal flow of the signal is inverted.
    
    \item \textbf{Time Warp} arbitrarily accelerates or decelerates the speed of signal. It helps overcome inter-subject speed variations, allowing the model to learn features which are robust to changes in temporal duration.
\end{itemize}

\subsubsection{Weak Augmentation} 
Weak augmentation serves as an anchor view that preserves the semantic integrity of the original data. To achieve this, we applied a combination of Jittering and Scaling to the original signal. The details of the two weak augmentation techniques
are as follows:

\begin{itemize}
    \item \textbf{Jittering} injects gaussian noise with a standard deviation of 0.01 into the original signal. It simulates inherent sensor noise.
    
    \item \textbf{Scaling} randomly scales the signal amplitude within a range of approximately 2\%. It addresses inter-subject magnitude variations, ensuring the model remains invariant to minor amplitude fluctuations.
\end{itemize}

\subsection{Evaluation Metrics}
We employed Balanced Accuracy (B-ACC) and Macro F1-Score (MF1) as performance metrics to deal with the class imbalance present in the datasets. The details of each metric are as follows:

\begin{itemize}
    \item \textbf{Balanced Accuracy (B-ACC)} is defined as the arithmetic mean of recall for each class. This metric is suitable for evaluating the model's overall discriminative capability in scenarios with severe data imbalance, ensuring that performance is not biased toward majority classes.
    \begin{equation}
        \text{B-ACC} = \frac{1}{C} \sum_{i=1}^{C} \text{Recall}_i
    \end{equation}

    \item \textbf{Macro F1-Score (MF1)} is defined as the unweighted mean of per-class F1-scores. This metric treats all classes equally, thereby fairly reflecting classification performance on minority classes.
    \begin{equation}
        \text{MF1} = \frac{1}{C} \sum_{i=1}^{C} \frac{2 \cdot \text{Precision}_i \cdot \text{Recall}_i}{\text{Precision}_i + \text{Recall}_i}
    \end{equation}
    Here, $C$ denotes the number of classes.
\end{itemize}

\section{RESULT}
\subsection{Comparison of Augmentation Strategies}
As presented in Table 1, the performance of single augmentation techniques varied significantly depending on the dataset characteristics. For Sleep-EDFX, \textit{Time Masking}, which preserves temporal context, demonstrated the most effective performance (B-ACC: 65.52\%; MF1: 61.83\%). 
For CHB-MIT, \textit{Crop \& Resize}, which emphasizes local patterns, achieved superior performance (B-ACC: 69.12\%, MF1: 64.25\%). In contrast, \textit{Time Warp}, which distorts temporal dynamics, yielded poor results on both datasets. This strategy recorded the lowest B-ACC of 62.49\% and 65.10\%, respectively. 

Consequently, the \textit{Random Selection} strategy revealed clear limitations, as it indiscriminately applies even detrimental augmentations without considering the data context. 
In contrast, RL-BioAug achieved remarkable performance on both Sleep-EDFX (B-ACC: 72.72\%, MF1: 69.55\%) and CHB-MIT (B-ACC: 76.24\%, MF1: 71.50\%). These results represent a substantial improvement of approximately 7.1\%p to 7.7\%p across both metrics compared to the best performing single baseline for each dataset. These findings validate that RL-BioAug adaptively selects optimal augmentations which preserve semantic integrity based on each sample.

\subsection{Ablation Experiments}
\subsubsection{Impact of Top-K Sampling Strategy}

Table ~\ref{tab:table2} summarizes the model performance with respect to the Top-$K$ sampling strategy. The model yielded the highest performance with $K=3$ on both Sleep-EDFX (B-ACC: 72.72\%, MF1: 69.55\%) and CHB-MIT (B-ACC: 76.24\%, MF1: 71.50\%). 
Specifically, when $K=1$, the selection was restricted to the single policy with the highest reward, thereby limiting the diversity of augmentations and resulting in relatively lower performance (e.g., Sleep-EDFX B-ACC: 71.04\%). Conversely, increasing $K$ to 4 or higher introduced unnecessary noise by incorporating sub-optimal policies with lower rewards into the training process (e.g., B-ACC: 71.23\% at $K=5$). These findings suggest that selecting the Top-3 policies represents the optimal balance, between ensuring augmentation diversity and maintaining training efficiency.

\begin{table}[t]
\centering
\begin{threeparttable}
\caption{Impact of Top-K Sampling Strategy.}
\label{tab:table2}
\small
\setlength{\tabcolsep}{12pt}
\renewcommand{\arraystretch}{1.2}
\begin{tabular}{c cccc}
\toprule
\multirow{2.5}{*}{\textit{Top-K}} & \multicolumn{2}{c}{Sleep-EDFX} & \multicolumn{2}{c}{CHB-MIT} \\
\cmidrule(lr){2-3} \cmidrule(lr){4-5}
 & B-ACC & MF1 & B-ACC & MF1 \\
\midrule
1                      & 71.04           & 68.18           & 74.85           & 69.82           \\
2                      & {\ul 71.35}    & 68.50           & 75.60           & 70.95           \\
\textbf{3}                      & \textbf{72.72} & \textbf{69.55} & \textbf{76.24} & \textbf{71.50} \\
4                      & 72.34           & {\ul 68.83}    & {\ul 75.92}    & {\ul 71.10}    \\
5                      & 71.23           & 68.32           & 75.15           & 70.45           \\ \bottomrule
\end{tabular}
\begin{tablenotes}
    \footnotesize
    \item - The \textbf{best results} in each column are shown in bold, while the \uline{second-best} results are underlined.
\end{tablenotes}
\end{threeparttable}
\end{table}

\subsubsection{Efficacy of Reward Mechanism} 

Table \ref{tab:table3} compares the model performance across different reward mechanisms. The results indicate that the Soft-KNN-based reward approach achieved the highest performance. 
This method ensured superior stability and enabled the agent to effectively converge on optimal policies by providing dense rewards. Consequently, it reached (B-ACC: 72.72\%, MF1: 69.55\%) on Sleep-EDFX and (B-ACC: 76.24\%, MF1: 71.50\%) on CHB-MIT, outperforming other reward mechanisms. In contrast, despite the advantage of operating without labels, the SSL loss approach tends to prefer trivial augmentations that are easy to learn. This led to the lowest performance (e.g., Sleep-EDFX B-ACC: 63.02\%). The accuracy-based method is limited in its ability to finely tune the policy network due to the sparsity of the reward signal (e.g., Sleep-EDFX B-ACC: 67.85\%).

\subsection{Analysis of Policy Networks} 

\begin{figure}[t]
    \centerline{\includegraphics[width=1.0\columnwidth]{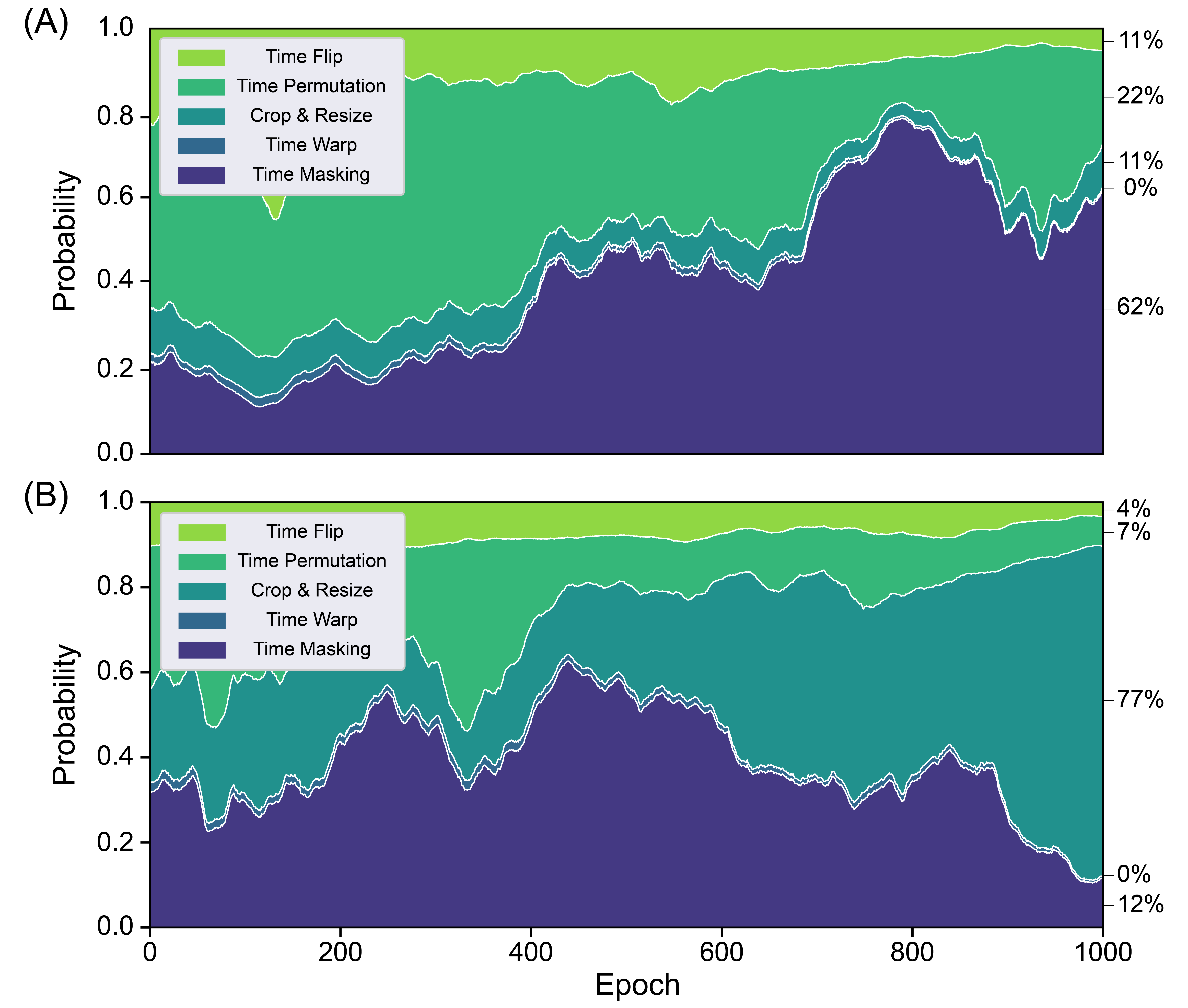}}
    \caption{Training dynamics of the data augmentation policy on (A) the Sleep-EDFX dataset and (B) the CHB-MIT dataset.}
    \label{fig:figure3}
\end{figure}

To validate the adaptability of the agent, we visualized the evolution of action probabilities during training (Fig.~\ref{fig:figure3}). The results confirm that the policy network autonomously converges toward optimal augmentation strategies tailored to the characteristics of each dataset.

Specifically, for the Sleep-EDFX (Fig.~\ref{fig:figure3}(A)), the selection probability of \textit{Time Masking} gradually increased, eventually  stabilizing at approximately 62\%. This aligns with the results in Table~\ref{tab:table1}, implying that the encoder's ability to infer global structure from the surrounding context is critical for sleep stage classification. In contrast, for the CHB-MIT (Fig.~\ref{fig:figure3}(B)), \textit{Crop \& Resize} was adopted as the dominant strategy, reaching a selection proabbility of 77\%. This corresponds with the domain knowledge that seizure detection relies on local waveform patterns rather than global structures \cite{shoeb2009application, niedermeyer2005electroencephalography}.

\begin{table}[t]
\centering
\begin{threeparttable}
\caption{Efficacy of Reward Mechanism on \\ Sleep-EDFX and CHB-MIT datasets.}
\label{tab:table3}
\small
\renewcommand{\arraystretch}{1.2}
\begin{tabular}{c cccc}
\toprule
\multirow{2.5}{*}{\textit{Reward}} & \multicolumn{2}{c}{Sleep-EDFX} & \multicolumn{2}{c}{CHB-MIT} \\
\cmidrule(lr){2-3} \cmidrule(lr){4-5}
 & B-ACC & MF1 & B-ACC & MF1 \\
\midrule
SSL Loss & 63.02 & 59.82 & 67.92 & 62.50 \\
Accuracy & \underline{67.85} & \underline{64.29} & \underline{72.15} & \underline{66.82} \\
\textbf{Soft-KNN (Ours)} & \textbf{72.72} & \textbf{69.55} & \textbf{76.24} & \textbf{71.50} \\
\bottomrule
\end{tabular}
\begin{tablenotes}
    \footnotesize
    \item - The \textbf{best results} in each column are shown in bold, while the \uline{second-best} results are underlined.
\end{tablenotes}
\end{threeparttable}
\end{table}

Collectively, these findings demonstrate that RL-BioAug is capable of autonomously identifying and applying optimal data augmentation techniques tailored to data characteristics, even without prior domain knowledge.

\section{DISCUSSION \& CONCLUSION}
The experimental results demonstrate that an adaptive strategy, actively exploring optimal transformations based on individual data characteristics, is essential for robust representation learning in EEG signals. This marks a significant departure from conventional static methods relying on expert heuristics \cite{eldele2021time}. Specifically, our analysis (Table~\ref{tab:table1}, Fig.~\ref{fig:figure3}) reveals that \textit{Time Masking} is the most effective strategy for the Sleep-EDFX dataset, where capturing long-range temporal context is crucial. In contrast, \textit{Crop \& Resize} yielded optimal performance for the CHB-MIT dataset, which requires capturing local anomalies such as seizures \cite{shoeb2009application, niedermeyer2005electroencephalography}.

In contrast, the \textit{Random Selection} strategy failed to match the optimal single-technique performance, as it applies augmentations stochastically without regard for intrinsic data characteristics. This underscores that effective augmentation strategies are highly task-dependent. RL-BioAug's superiority stems from the agent's ability to autonomously learn \textbf{safe and valid transformations}. By preserving signal integrity, it effectively eliminates the uncertainty associated with subjective human intervention.

Despite these contributions, our study presents several limitations. First, the agent's action space is confined to a pre-defined discrete pool. While effective at selecting combinations, the model cannot yet create novel transformations tailored to the signal's inherent characteristics. Second, the computational overhead inherent to RL imposes constraints on real-time deployment, particularly in resource-limited environments. Third, the reward mechanism relies on partial supervision. Given the prevalence of label noise in bio-signals \cite{karimi2020deep}, dependence on potentially imperfect ground truth risks introducing negative bias into the policy learning process.

Future research will extend validation beyond EEG to diverse bio-signals such as electrocardiograms (ECG) and electromyograms (EMG). Furthermore, we aim to advance beyond pre-defined discrete pools by establishing a dynamic generation pipeline. By exploring continuous action spaces and leveraging generative networks, our goal is to autonomously synthesize transformations that adapt to the signal's characteristics.

In conclusion, RL-BioAug achieved superior performance on both the Sleep-EDFX and CHB-MIT datasets by autonomously learning optimal data augmentation policies that preserve semantic structure without expert intervention. These results underscore the framework's robust generalizability across diverse pathological conditions and clinical tasks. Ultimately, this work establishes a critical foundation for extending unsupervised medical data analysis to label-free environments.

\addtolength{\textheight}{-12cm}   




\section*{ACKNOWLEDGMENT}


\bibliographystyle{IEEEtran}
\bibliography{reference}

\end{document}